\PassOptionsToPackage{table}{xcolor}
\PassOptionsToPackage{hyphens}{url}
\documentclass{alaya}
\usepackage{makecell}
\usepackage{url}
\usepackage{adjustbox}
\usepackage[sc]{mathpazo}
\usepackage{amsmath,amssymb}
\usepackage[numbers]{natbib}
\usepackage{array}

\newif\ifshowrevisions
\showrevisionsfalse
\definecolor{revisionred}{HTML}{BD2026}

\newcommand{\tablestyle}[2]{\setlength{\tabcolsep}{#1}\renewcommand{\arraystretch}{#2}\centering\footnotesize}
\renewcommand{\paragraph}[1]{\vspace{1.25mm}\noindent\textbf{#1}}

\title{MeshAxiom: An Agent-Native 3D Foundation Stack for Programmable Worlds}
\author[1]{Yonghao Yu}
\author[1]{Ruihan Yu}
\author[1]{Sho Kuno}
\author[1]{Muyao Niu}
\author[1]{Yu-Ju Tsai}
\author[1]{Zheng-Hui Huang}
\author[1]{Fengbo Lan}
\author[1]{Guixu Lin}
\author[1]{Runyi Li}
\author[1]{Chuanhao Li}
\author{Ming-Hsuan Yang}
\author[1,\dagger]{Kaipeng Zhang}
\author[1,\dagger]{Zhixiang Wang}

\affiliation[1]{Alaya Lab}

\abstract{
Agents that build 3D worlds need assets they can inspect, edit and reuse through code, yet dense generators such as TRELLIS.2 and Pixal3D produce hundreds of thousands to millions of triangles from one image.
We present \textbf{MeshAxiom}, an agent-native 3D foundation stack for building programmable worlds from explicit low-polygon meshes.
Its geometry backend, MeshAxiom-Base, generates vertices and triangles from a single image and a target face count.
We train it on 600,000 real-world Unreal Engine (UE) assets, each with fewer than 20,000 faces.
On a development set of 500 objects, MeshAxiom-Base at 2K, 4K and 8K requested faces obtains lower Chamfer, Hausdorff and edge distances and higher ULIP similarity than TRELLIS, TRELLIS.2 and Pixal3D after their official simplification to 8K faces.
Requests of 2K, 4K and 8K faces yield 1,978, 4,188 and 8,837 faces on average. At a budget of 4K faces, generation takes a median of 6.7\,s on one NVIDIA H200 GPU.
In our export, rigging and assembly tests, the agent used 2\% to 32\% fewer language model tokens and took 7\% to 76\% less time with low-polygon inputs than with dense inputs.
The agent checks geometry, makes bounded repairs to mesh files and calls external tools to refine materials. It records accepted assets for reuse in assembly and scene programs.
In a pilot of skill-level recursive self-improvement (RSI) with fixed model weights, a workflow learned from episode traces used fewer tokens and took less time per successful asset than the original workflow under the same quality checks.
We use the stack to build 123 assets, 19 assemblies and 12 animated scenes.
The asset library, inference code and model weights are publicly available.
}

\project{\url{https://alaya-lab.github.io/MeshAxiom/}}
\code{\url{https://github.com/AlayaLab/MeshAxiom}}
\correspondence{Zhixiang Wang (Project Lead) and Kaipeng Zhang}
\date{November 26, 2026}

\begin{document}
\maketitle
\begin{figure}[t]
  \centering
  \includegraphics[width=\linewidth]{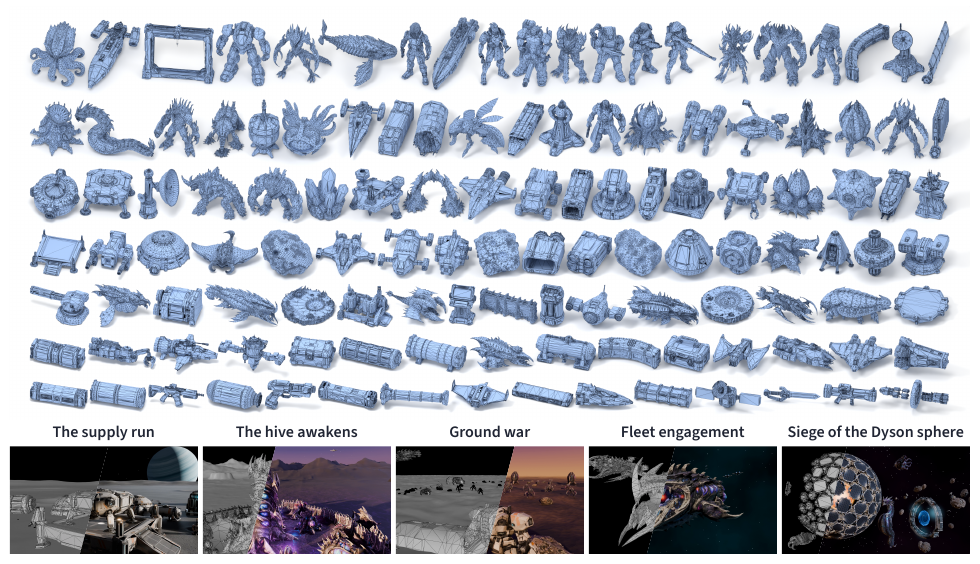}
  \caption{\textbf{MeshAxiom assets and animated worlds.} The upper panel shows all 123 library assets in one scene with blue shading and visible triangle edges. Below are five of the twelve animated scenes. Each diagonal split shows the same frame, camera and meshes, with geometry on the left and materials on the right.}
  \label{fig:teaser}
\end{figure}

\section{Introduction}
A programmable world is one that an agent builds and edits through tools and code that act on explicit scene state, such as meshes and object transforms.
Such a world needs 3D assets that the agent can inspect, modify and reuse, and current approaches follow one of three routes.
First, an agent can write a program to maintain scene state and use a video model to render its appearance from a coarse proxy~\cite{chen2026codeworldmodel}. Geometric editing still requires access to explicit meshes.
Second, agents can write Blender scripts to assemble scenes~\cite{hu2024scenecraft}, edit graphics programs using rendered feedback~\cite{huang2024blenderalchemy}, or control procedural modeling tools~\cite{sun2023threedgpt}.
Third, the agent calls a 3D generation model.
Dense image-to-3D generators~\cite{li2025triposg,xiang2025trellis2,xiang2024trellis,zhao2025hunyuan3d} produce detailed meshes whose large geometry arrays make inspection, editing and rendering more expensive. The dense models we evaluate generate hundreds of thousands to millions of triangles.
Commercial services such as Tripo P2 and Meshy T2 return compact triangle meshes directly from images through hosted APIs~\cite{meshyT2Api2026,tripoP2Api2026}, while open models for generating artist low-poly meshes~\cite{chen2024meshanythingv2,lionar2025treemeshgpt,zhao2025deepmesh} usually take a point cloud as input.

We train a model with open weights that generates explicit low-polygon meshes from images. Around it we build an agent harness and workflows for constructing scenes.
Figure~\ref{fig:teaser} shows the resulting asset library and frames from five animated scenes built from its meshes.
Explicit low-polygon meshes give agents a stable interface between an instruction and the resulting scene: their vertices and triangles can be checked by tools, revised as files and reused across scene programs.
MeshAxiom is a 3D foundation stack organized around these shared mesh assets.
We call it \emph{agent-native} because agents use it through tools that act on asset files: they receive compact summaries and rendered views, while the files retain the geometry and generation recipe for later operations.
Figure~\ref{fig:workflow} follows a Dyson panel from its reference image to the assembled sphere. It also shows how a learner proposes a workflow change from episode traces and how the change is tested on new objects.

MeshAxiom-Base converts one image and a requested face budget into explicit triangles. Following Nexus~\cite{wang2026nexus}, it generates vertex positions and connectivity in separate stages.
Figure~\ref{fig:base_model} shows how a vertex diffusion transformer generates vertices as occupied octree cells, from coarse to fine on a $512^3$ grid. A second transformer generates a set of latents, one per vertex. A topology autoencoder decodes these latents into connectivity.
Budget conditioning lets the caller request geometry at different complexities before assembly or export.
The agent can inspect triangle arrays, component counts and rendered views, request small repairs and call external tools for texturing.
A generation run records the reference image, face budget and seed. The asset library links meshes to material variants, and scene layouts place each instance by asset key and transform.

We compare 14 methods on 500 development objects that are disjoint from the training data.
Table~\ref{tab:image500} shows that at 2K, 4K and 8K requested faces, MeshAxiom-Base has lower mean Chamfer, Hausdorff and edge Chamfer distances than every baseline.
The baselines include TRELLIS~\cite{xiang2024trellis}, TRELLIS.2~\cite{xiang2025trellis2} and Pixal3D~\cite{li2026pixal3d} after their official simplification to 8K faces.
MeshAxiom-Base also has the highest ULIP~\cite{xue2022ulip} similarity.
Its output face counts follow the requested budget. At a budget of 4K faces, median generation time is 6.7\,s.

We also use the stack to build a library of 123 assets that span mechanical, organic and environmental forms, and we reuse them in 19 assemblies and 12 animated scenes.
In a pilot study of skill-level recursive self-improvement (RSI), a learner reads traces of earlier episodes and proposes a reusable workflow skill. The candidate uses fewer language model tokens and less execution time per successful asset than the original workflow.
The asset library, inference code and model weights are publicly available. We plan to release the training code and the remaining workflow tools.

We make three contributions.
\begin{itemize}\interlinepenalty=10000 \csname @beginparpenalty\endcsname=10000
  \item We release the weights of MeshAxiom-Base, a model built on the Nexus framework that generates explicit low-polygon meshes from one image and a face budget.
  \item We build an agent harness with tools for checking meshes, repairing geometry and adding materials, and an interface for learning workflow skills. Assembly and scene programs reuse the resulting asset records. We use the stack to build 123 assets, 19 assemblies and 12 animated scenes.
  \item We evaluate MeshAxiom-Base against 13 other methods on 500 development objects. It has the lowest Chamfer, Hausdorff and edge distances at all three face budgets. We also measure budget control, inference time and downstream agent effort.
\end{itemize}

\section{Related Work}
\paragraph{World generation and programmable state.}
Video world models such as GameNGen~\cite{valevski2024gamengen}, Genie~\cite{bruce2024genie} and Matrix-Game~\cite{zhang2025matrixgame} generate interactive environments frame by frame. Diffusion as Shader~\cite{gu2025das} conditions video appearance on explicit 3D tracking videos. These methods produce rendered frames. Editing individual assets requires access to their geometry.
Other systems instead keep the world state in executable code.
In Code World Model~\cite{chen2026codeworldmodel} and Programmable World Model~\cite{huang2026pwm}, a coding agent writes programs that define the world's state and rules, and a video model renders observations from them.
StreamRender-H3~\cite{lu2026streamrenderh3}, EdenLive~\cite{huang2026edenlive} and Matrix-Game~4.0~\cite{zhang2026matrixgame4} similarly pair a code-driven engine or agent with a real-time video renderer.
The renderer's inputs include projected 3D bounding boxes in Programmable World Model, semantic maps in StreamRender-H3 and reference video in EdenLive.
Recursive Code World Models~\cite{li2026recursivecodeworld} reconstruct complex 3D scenes from a single image as executable scene programs.
In MeshAxiom, the scene state contains explicit low-polygon meshes generated from images. Tools inspect and repair these meshes, and scene programs instantiate and transform them. The scenes in our demonstrations are engine renders of these meshes.

\paragraph{Sparse 3D diffusion.}
XCube~\cite{ren2024xcube} generates sparse voxel hierarchies by applying latent diffusion from coarse to fine, and OctFusion~\cite{xiong2025octfusion} combines octree latents with diffusion at multiple scales.
TRELLIS~\cite{xiang2024trellis} first generates a sparse spatial structure and then the structured latents attached to it.
Nexus~\cite{wang2026nexus} applies the sparse hierarchy to meshes directly: the occupied cells at the finest octree level are the mesh vertices, and a separate topology model predicts their connectivity.
MeshAxiom-Base uses this formulation to generate low-polygon meshes from images. We release the model weights and build tools that let agents work with its output.

\paragraph{Explicit mesh generation.}
TRELLIS~\cite{xiang2024trellis}, TRELLIS.2~\cite{xiang2025trellis2}, Pixal3D~\cite{li2026pixal3d} and TripoSG~\cite{li2025triposg} generate detailed surfaces from structured or implicit representations.
Models for generating artist meshes predict vertices and faces directly.
PolyGen~\cite{nash2020polygen} generates them sequentially and supports image conditioning.
Later methods include MeshGPT~\cite{siddiqui2024meshgpt}, MeshAnything V2~\cite{chen2024meshanythingv2}, EdgeRunner~\cite{tang2024edgerunner}, BPT~\cite{weng2025scaling}, TreeMeshGPT~\cite{lionar2025treemeshgpt}, DeepMesh~\cite{zhao2025deepmesh}, Mesh-Silksong~\cite{song2026silksong}, MeshRipple~\cite{lin2026meshripple} and FastMesh~\cite{kim2025fastmesh}.
MeshFlow~\cite{li2026meshflow} models compact topology in a continuous representation.
Like Nexus, LATO.2~\cite{long2026lato2} separates vertex generation from topology generation. Its vertex flow is conditioned on the image and allows control of the vertex count. Its topology flow is conditioned on the generated vertices.
The vertex stage samples structured latents that are decoded into positions. Our model predicts occupied octree cells level by level, conditioned on the requested face count.
Because the LATO.2 pipeline we evaluate starts from a reconstruction, we group it with the methods conditioned on geometry.

\paragraph{Agents and 3D tools.}
Authoring agents such as 3D-GPT~\cite{sun2023threedgpt}, SceneCraft~\cite{hu2024scenecraft} and BlenderAlchemy~\cite{huang2024blenderalchemy} build or edit 3D content by writing Blender code, which motivates tool interfaces for editing explicit assets and scenes.
MeshAxiom gives authoring agents a mesh generator together with asset records, checks and scene programs, and our case studies measure how low-polygon geometry affects downstream agent work.

\section{Method}
\label{sec:method}
\begin{figure}[t]
  \centering
  {\color[HTML]{202533}\includegraphics[width=\linewidth]{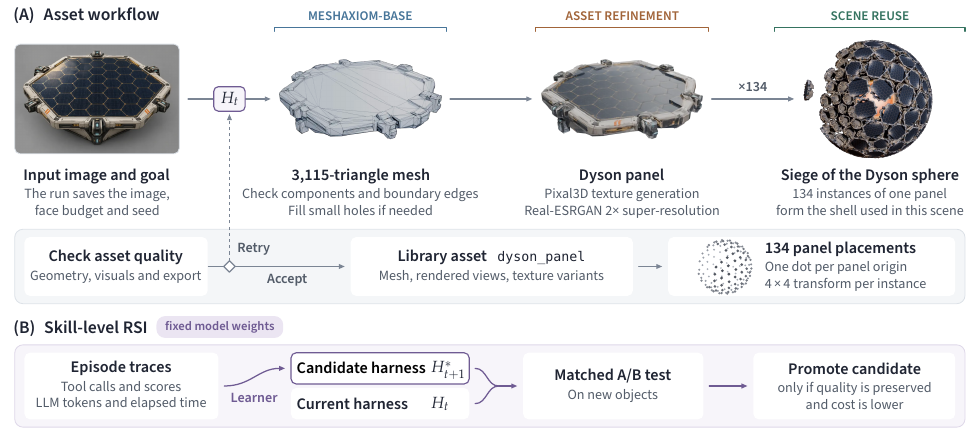}}
  \caption{\textbf{Asset workflow and skill-level recursive self-improvement (RSI).} (A) The agent calls MeshAxiom-Base to generate a 3,115-triangle Dyson panel and Pixal3D to produce its initial texture. Once the panel passes geometry, appearance and export checks, 134 instances form the shell around a star in \emph{Siege of the Dyson sphere}. (B) With model weights fixed, a learner uses episode traces to propose a candidate harness $H^{*}_{t+1}$. The candidate replaces $H_t$ only if it preserves quality and lowers execution cost in matched A/B tests on new objects.}
  \label{fig:workflow}
\end{figure}

Figure~\ref{fig:workflow} shows how MeshAxiom's generator, agent and scene programs work with the same asset files.
The agent is a large language model (LLM) whose harness $H_t$ comprises its prompts, skills, memory and tool routes.
A task gives it an input image, a goal and a face budget.
It calls MeshAxiom-Base to generate vertices and connectivity, then runs geometry and material tools on the resulting mesh.
The agent checks the asset's geometry, appearance and exportability. If the asset passes these checks, the agent adds it to the library for reuse in assembly and scene programs. If a check fails, the agent revises and retries the task through the same harness.
The episode trace records token usage and elapsed time for each run. We sum the language model's uncached input tokens and its output tokens. These include the tool results it reads and the calls it writes.
MeshAxiom-Base can also run directly from a script. In the agent workflow, its GPU runtime contributes to total elapsed time.
A learner can use these traces to propose changes to the harness.

\subsection{MeshAxiom-Base: image-conditioned mesh generation}
\label{sec:base}
\begin{figure}[t]
  \centering
  \includegraphics[width=\linewidth]{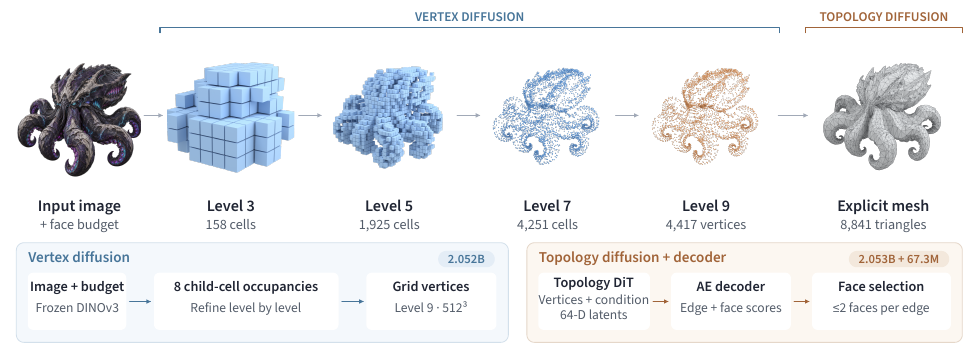}
  \caption{\textbf{MeshAxiom-Base generates explicit triangles from an image and a face budget.} The cell views group the 4,417 vertices of this Star Kraken example into occupied octree cells at successive resolutions, reaching a $512^3$ grid at level 9. Topology diffusion samples one latent per vertex. The autoencoder (AE) decoder scores candidate edges and faces. The resulting mesh has 8,841 triangles. Parameter counts exclude the frozen DINOv3 image encoder. The topology stage lists the diffusion model and the AE decoder separately.}
  \label{fig:base_model}
\end{figure}

MeshAxiom-Base generates a low-polygon triangle mesh from one image and a requested face budget.
Figure~\ref{fig:base_model} illustrates the decoupled vertex and topology diffusion framework adapted from Nexus~\cite{wang2026nexus}. One model generates vertex positions and a second generates their connectivity.

\paragraph{Representation.}
We center each training mesh on its bounding box, scale it so that its longest side is 1.9 inside $[-1,1]^3$, and quantize its vertices to a $512^3$ grid whose cells are the leaves of a nine-level octree.
Each occupied cell at level 9 is one mesh vertex, so the octree stores a sparse point set rather than a filled voxel volume.

\paragraph{Vertex diffusion.}
The vertex model generates the octree from coarse to fine.
At each level it predicts, for every occupied cell, the occupancies of its eight child cells, and the children predicted as occupied become the cells of the next level.
The model is conditioned on image tokens from a frozen DINOv3 encoder~\cite{simeoni2025dinov3} and a token for the face budget.
To form this token, we encode the requested face count $F_{\mathrm{req}}$ as $\log(1+F_{\mathrm{req}})/\log F_{\max}$, where $F_{\max}=20{,}000$ is the face limit of our training data.
A two-layer MLP maps this scalar to a token that is appended to the image context.
The face budget is a soft, learned condition, so the generated face count can differ from the request.

\paragraph{Topology diffusion.}
Connectivity is encoded by a topology autoencoder as one 64-dimensional latent per vertex.
Its decoder first scores candidate edges and then scores cycles of three edges that form candidate triangles.
Given the generated vertices and the image, the topology model samples these latents.

\paragraph{Training.}
We train both transformers to predict velocity with flow matching~\cite{liu2022rectified}.
Let $x_1$ be the clean target and $x_0\sim\mathcal{N}(0,I)$ be Gaussian noise. The targets are child cell occupancies for the vertex model and one latent per vertex for the topology model.
With $x_t=(1-t)\,x_0+t\,x_1$, the network $v_\theta$ regresses the velocity of this path under a squared error,
\begin{equation}
  \mathcal{L}=\mathbb{E}_{t,\,x_0,\,x_1}\bigl\lVert v_\theta(x_t,t,c)-(x_1-x_0)\bigr\rVert_2^2 ,
  \label{eq:flow}
\end{equation}
where $c$ denotes the conditioning input. At inference, vertex diffusion uses the image tokens and the token for the face budget. Topology diffusion uses the generated vertices and the image.
We train the topology autoencoder beforehand using binary cross-entropy losses on edge and face indicators and a KL regularizer. We then freeze its encoder and use its standardized vertex latents as the targets for topology diffusion.

\paragraph{Architecture and data.}
The vertex and topology transformers each have 22 blocks of width 2,048. They have 2.052B and 2.053B trainable parameters, respectively, excluding the frozen DINOv3 encoder. The topology autoencoder has 111.512M parameters.
For training, we curate 600,000 real-world Unreal Engine (UE) assets, each with fewer than 20,000 faces.
Mesh targets use the normalization and quantization above, and rendered images supply the visual condition.
Our development set consists of 500 additional UE assets. We verified that none occurs in the training data.

\paragraph{Inference.}
Unless stated otherwise, each octree level is sampled with 20 solver steps.
The released recipe uses DPM-Solver++~\cite{lu2022dpmsolverpp} and classifier-free guidance~\cite{ho2022cfg}, with a guidance scale of 3 for the first four levels and 5 for levels 5 through 9.
Since later levels only refine the cells that the coarse levels keep, we draw $K=8$ candidates for the first three levels and keep the one that agrees best with the others,
\begin{equation}
  k^{\star}=\arg\max_{k}\;\frac{1}{K-1}\sum_{j\neq k}\mathrm{IoU}\bigl(S_k,S_j\bigr),
  \label{eq:mbr}
\end{equation}
where $S_k$ is the set of cells occupied by candidate $k$ at level 3. Only candidate $k^{\star}$ is refined to level 9.
We denote minimum Bayes risk (MBR) selection with $K=8$ by MBR8.
Topology sampling uses 20 steps, a guidance scale of 3 and an initial noise scale of 0.5.
We decode two sets of latents: the sampled ones, and an estimate of their conditional mean that the network gives at $t=0$, the noise end of the path in Equation~\eqref{eq:flow}.
Faces decoded from the sampled latents are kept first, and a face decoded from the mean estimate is added only if it shares an edge with a face already kept.
Throughout, no edge may have more than two incident faces.
This limit excludes non-manifold edges.
We call the resulting mesh the \emph{raw output} of the model.

\subsection{Asset records and tools}
\label{sec:refine}
The generation run saves the reference image and records the requested face budget and random seed in \texttt{request.json}.
In the library, each \emph{asset record} links the mesh to its reference image and to gray and material renders. The material catalog links texture variants to the same mesh.

Geometry tools read the vertex and triangle arrays to count connected components and boundary edges. The agent can inspect rendered views and discard a repair that fails a quality check.
For display, the tools can fill small holes when needed and prepare normals for each connected component. These operations act on copies. The geometry benchmark and face-budget sweep use the raw model output.

We generate the initial texture with Pixal3D~\cite{li2026pixal3d}, then project the reference image onto visible surfaces and repaint multiple views using renders of the untextured mesh as guides.
For high-resolution variants, we use Real-ESRGAN x2plus~\cite{wang2021realesrgan} to double the width and height of the base color texture.

\subsection{Scene programs and skill-level RSI}
\label{sec:modules}
Accepted assets become \emph{modules} that programs can reference.
An assembly program instantiates modules and moves them through explicit transforms while leaving their geometry and materials unchanged.
The Dyson sphere layout uses 134 instances of the library mesh \texttt{dyson\_panel} to form the shell around the star. Each panel has its own $4\times4$ transform. The dots in panel~A of Figure~\ref{fig:workflow} show where those transforms place the panel origins.
A scene program adds terrain placement, rigs, motion, cameras, effects and event timing.
Because the scene state is explicit, any frame can be rendered either as untextured geometry or with materials from the same state and camera.
An optional video model can regenerate the appearance of an engine render from prompts and references, with the engine render kept alongside it.

In panel~B of Figure~\ref{fig:workflow}, the learner reads episode traces containing assets, scores, tool calls, token counts and elapsed times. It revises prompts, memory or tool choices to produce a candidate harness $H^{*}_{t+1}$ with model weights fixed. We compare the current and candidate harnesses on the same new objects and adopt the candidate only if it preserves quality at lower execution cost.

\section{Experiments}
The experiments assess mesh quality, compare actual and requested face counts, and measure inference time.
Assembly and scene examples illustrate asset reuse, and agent studies quantify the effort required for downstream tasks and the cost of running a learned workflow.

\subsection{Geometry quality and complexity}
\label{sec:image500}
\paragraph{Setup.}
We evaluate 19 configurations of 14 methods on 500 UE objects held out from generator training.
Each comparison starts from the same $512\times512$ image of the object. The reference mesh uses the same $512^3$ grid resolution as the training meshes.
The methods in Panel~A of Table~\ref{tab:image500} take the image directly.
The methods in Panel~B, including the evaluated LATO.2 pipeline, are conditioned on geometry: they receive frozen TRELLIS.2~\cite{xiang2025trellis2} outputs for the same image, converted to their input format, so their scores include upstream errors.
No method is given ground-truth geometry.

\begin{table}[t]
  \centering
  \tablestyle{3pt}{1.12}
  \caption{\textbf{Geometry quality and complexity on 500 development objects.}
    Each of the 19 configurations produced a mesh for every object.
    The Request column gives the requested face count, a soft target of 2,600 vertices (2.6K-V), no count condition (Auto), or official simplification toward 8,000 triangles ($8\mathrm{K}^{\ast}$).
    V and F are the mean vertex and face counts.
    We use the same winding propagation for all methods before computing NC. Bold marks the best unrounded mean for each metric.}
  \label{tab:image500}
  \begin{adjustbox}{max width=\linewidth}
  \newlength{\imageMetricWidth}
  \settowidth{\imageMetricWidth}{Uni3D-Sim$\uparrow$}
  \addtolength{\imageMetricWidth}{2pt}
  \begin{tabular}{lc*{8}{>{\centering\arraybackslash}p{\imageMetricWidth}}}
    \toprule
    Method & Request & CD$\downarrow$ & HD$\downarrow$ & ULIP-Sim$\uparrow$ & Uni3D-Sim$\uparrow$ & ECD$\downarrow$ & NC$\downarrow$ & V & F \\
    \midrule
    \multicolumn{10}{l}{\textit{A: Direct image-conditioned generation}} \\
    TRELLIS~\cite{xiang2024trellis} & Auto & 0.3009 & 0.5948 & 0.1392 & 0.3462 & 0.3114 & {\bfseries 0.0143} & 223,984 & 447,964 \\
    TRELLIS~\cite{xiang2024trellis} & 8K$^\ast$ & 0.3010 & 0.5946 & 0.1383 & 0.3461 & 0.3137 & 0.1564 & 3,986 & 8,000 \\
    TRELLIS.2~\cite{xiang2025trellis2} & Auto & 0.3193 & 0.6273 & 0.1413 & 0.3460 & 0.3303 & 0.0157 & 3,192,749 & 6,525,556 \\
    TRELLIS.2~\cite{xiang2025trellis2} & 8K$^\ast$ & 0.3237 & 0.6249 & 0.1495 & 0.3409 & 0.3346 & 0.2102 & 2,747 & 7,801 \\
    Pixal3D~\cite{li2026pixal3d} & Auto & 0.3291 & 0.6630 & 0.1422 & {\bfseries 0.3501} & 0.3411 & 0.0196 & 2,700,134 & 5,798,238 \\
    Pixal3D~\cite{li2026pixal3d} & 8K$^\ast$ & 0.3346 & 0.6621 & 0.1520 & 0.3424 & 0.3491 & 0.2203 & 2,227 & 7,789 \\
    EdgeRunner~\cite{tang2024edgerunner} & 4K & 0.4739 & 0.9114 & 0.1135 & 0.1956 & 0.4924 & 0.4157 & 1,904 & 3,734 \\
    \addlinespace[2pt]
    {\bfseries MeshAxiom-Base (Ours)} & 2K & 0.1278 & 0.3119 & {\bfseries 0.1602} & 0.3422 & 0.1326 & 0.2732 & 1,038 & 1,978 \\
    {\bfseries MeshAxiom-Base (Ours)} & 4K & 0.1250 & {\bfseries 0.3053} & 0.1583 & 0.3463 & 0.1286 & 0.2532 & 2,162 & 4,188 \\
    {\bfseries MeshAxiom-Base (Ours)} & 8K & {\bfseries 0.1249} & 0.3078 & 0.1557 & 0.3490 & {\bfseries 0.1286} & 0.2436 & 4,545 & 8,837 \\
    \midrule
    \multicolumn{10}{l}{\textit{B: Same-image TRELLIS.2 geometry-conditioned pipelines}} \\
    MeshFlow~\cite{li2026meshflow} & Auto & 0.3242 & 0.6294 & 0.1462 & 0.3220 & 0.3381 & 0.3062 & 965 & 2,971 \\
    MeshAnything V2~\cite{chen2024meshanythingv2} & Auto & 0.3865 & 0.7196 & 0.1318 & 0.2573 & 0.3988 & 0.2615 & 960 & 1,636 \\
    \addlinespace[2pt]
    BPT~\cite{weng2025scaling} & Auto & 0.5708 & 0.9933 & 0.1017 & 0.1884 & 0.5881 & 0.3917 & 816 & 1,412 \\
    TreeMeshGPT~\cite{lionar2025treemeshgpt} & Auto & 0.5809 & 1.0784 & 0.0953 & 0.1682 & 0.6009 & 0.1260 & 5,309 & 9,799 \\
    MeshRipple~\cite{lin2026meshripple} & Auto & 0.3910 & 0.7598 & 0.0985 & 0.2303 & 0.4071 & 0.4582 & 3,929 & 7,451 \\
    DeepMesh~\cite{zhao2025deepmesh} & Auto & 0.5005 & 0.9634 & 0.0807 & 0.1679 & 0.5121 & 0.1932 & 10,049 & 16,487 \\
    Mesh-Silksong~\cite{song2026silksong} & Auto & 0.3548 & 0.6921 & 0.1419 & 0.2819 & 0.3658 & 0.2094 & 2,173 & 3,925 \\
    FastMesh~\cite{kim2025fastmesh} & Auto & 0.3402 & 0.6651 & 0.1323 & 0.2826 & 0.3548 & 0.4719 & 3,810 & 44,733 \\
    LATO.2~\cite{long2026lato2} & 2.6K-V & 0.3209 & 0.6270 & 0.1342 & 0.3333 & 0.3323 & 0.4723 & 2,642 & 8,265 \\
    \bottomrule
  \end{tabular}
  \end{adjustbox}
\end{table}

\paragraph{Face budgets.}
Our three configurations request 2K, 4K and 8K faces with the same weights, 20 sampling steps and MBR8, and none uses ground-truth face counts.
External methods use their native 8K setting if available, otherwise 4K, then 2K, then their highest released tier.
For the three dense generators we report both the native output and the output of the method's official simplifier run toward 8,000 triangles, which we call official simplification.
A model's maximum output capacity is not treated as a request, so models with only a capacity limit are listed as Auto.
LATO.2 uses a soft target of 2,600 vertices and produces 8,265 faces on average.

\paragraph{Metrics.}
For each object, we evaluate all methods in the same ground-truth frame, in which the longest side of the ground-truth bounding box spans two units. Each output is mapped into this frame with its method's fixed coordinate convention.
We sample 50K surface points per mesh in proportion to face area and compute Euclidean nearest neighbor distances in both directions. Chamfer distance (CD) sums the directional means, and Hausdorff distance (HD) takes the overall maximum.
Edge Chamfer distance (ECD) applies CD to 50K points per mesh sampled from unique triangle edges in proportion to length, capturing both shape and tessellation.
Semantic scores are cosine similarities between embeddings of the mesh point cloud and the input image composited on black. We use 8,192 points for ULIP~\cite{xue2022ulip} with Point-BERT~\cite{yu2022pointbert} and 10,000 for Uni3D-g~\cite{zhou2023uni3d} with EVA-CLIP~\cite{sun2023evaclip}.
Only these point clouds use unit-ball normalization and a fixed Z-up to Y-up rotation. Uni3D receives constant point colors.
Normal consistency (NC) averages $1-\mathbf n_i^\top\mathbf n_j$ over nondegenerate face pairs sharing an edge with exactly two incident faces. Lower values indicate closer agreement between adjacent face normals within the same mesh.
Before computing NC, we propagate face winding deterministically across these edges for every mesh, reordering triangle vertices.

\paragraph{Results.}
Across the three face budgets in Table~\ref{tab:image500}, MeshAxiom-Base has lower mean CD, HD and ECD than every baseline.
With a 2K budget, its CD is 0.1278, less than half the best baseline value of 0.3009 for native TRELLIS. Raising the budget to 8K more than quadruples the mean face count, while CD falls to 0.1249.
LATO.2 and MeshFlow have the lowest CD values in Panel~B, both slightly above the mean CD of their TRELLIS.2 inputs.
The three dense generators each average over 400K faces per mesh. Their official simplifiers reduce the mean face counts to about 8K, changing CD by less than 0.006 in every case.
Simplification raises each generator's NC to more than ten times its native value. All three still have lower NC than MeshAxiom-Base at the 8K budget.
Edge placement may contribute to this gap because adjacent face normals can differ sharply along a crease even when the surface matches the reference.
All three MeshAxiom-Base configurations score above every baseline on ULIP. On Uni3D, our 8K configuration ranks second at 0.3490, just behind native Pixal3D at 0.3501.

\begin{figure}[!t]
  \centering
  \includegraphics[width=\linewidth]{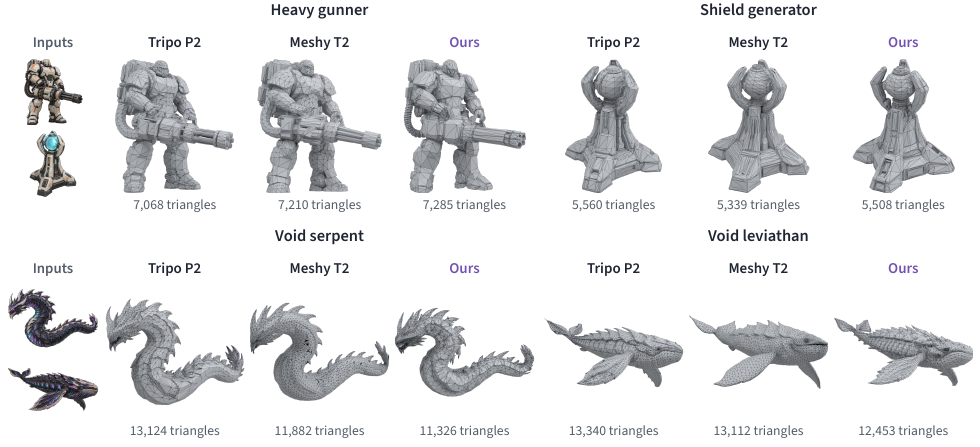}
  \caption{\textbf{Comparison with commercial models at similar triangle counts.} In each row, the upper and lower inputs at left correspond to the first and second comparison groups. All meshes share neutral shading, edge overlays and framing, with cameras estimated from the input views. Labels give triangle counts. All meshes are shown without decimation.}
  \label{fig:commercial}

  \par\addvspace{\floatsep}
  \nextfloat
  \includegraphics[width=\linewidth]{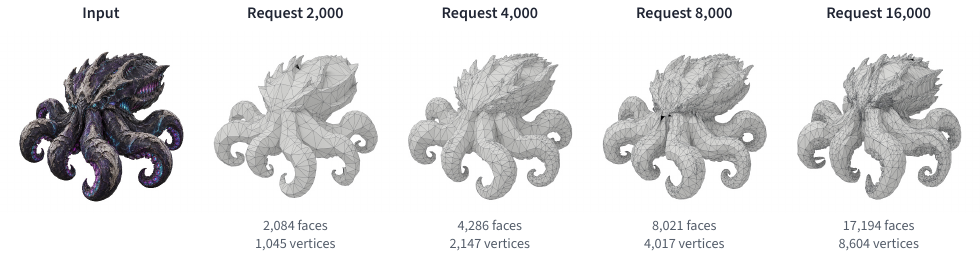}
  \caption{\textbf{Native Star Kraken meshes at four face budgets.} MeshAxiom-Base generates each mesh independently. All four are rendered with the same camera, lighting and display scale. Labels give the face and vertex counts of each output.}
  \label{fig:budget}
\end{figure}

\subsection{Qualitative comparison with commercial models}
\label{sec:commercial}
Figure~\ref{fig:commercial} compares MeshAxiom-Base with Tripo P2~\cite{tripoP2Api2026} and Meshy T2~\cite{meshyT2Api2026} on four assets at comparable output triangle counts.
All three methods used the same $512\times512$ image per asset. We gave the two APIs the same face target and made one untextured call to each per asset, using P2's triangle output and T2's Smart Topology.

\subsection{Control of face count and inference speed}
\label{sec:budget}\label{sec:inference}
Figure~\ref{fig:budget} shows the Star Kraken at 2K, 4K, 8K and 16K requested faces. All four runs use the same image, weights and seeds, with MBR8 and 20 sampling steps.
For requests of 2K, 4K and 8K faces, Table~\ref{tab:image500} gives mean outputs of 1,978, 4,188 and 8,837 faces on the development set, deviations of $-1.1\%$, $+4.7\%$ and $+10.5\%$, respectively.

We time MBR8 candidate selection, topology sampling and decoding with fixed vertex and topology checkpoints, 20 sampling steps and no distillation. We batch the eight coarse vertex candidates, cache cross-attention, vectorize postprocessing and use CUDA graphs~\cite{nvidia2026cudagraphs} to reduce kernel launch overhead.
The native face counts in Table~\ref{tab:inference} exceed their targets by 0.3\% to 7.5\%. None of the four meshes has a non-manifold edge. Generation takes 5.2 to 21.8\,s. Across all 500 development objects, median geometry generation time at the 4K target is 6.7\,s on one NVIDIA H200 GPU.

\begin{table}[!t]
  \centering
  \tablestyle{5pt}{1.1}
  \caption{\textbf{Budget control and inference time on Star Kraken.} Boundary counts open edges and Components counts connected components among the vertices used by the mesh. No run reaches a capacity limit. Times cover sampling and decoding on one NVIDIA H200 GPU with both models resident, excluding loading.}
  \label{tab:inference}
  \begin{tabular}{rrrrrr}
    \toprule
    Request & Faces & Vertices & Boundary & Components & Time (s) \\
    \midrule
    2,000 & 2,084 & 1,045 & 10 & 1 & 5.2 \\
    4,000 & 4,286 & 2,147 & 0 & 3 & 7.5 \\
    8,000 & 8,021 & 4,017 & 29 & 2 & 11.5 \\
    16,000 & 17,194 & 8,604 & 38 & 3 & 21.8 \\
    \bottomrule
  \end{tabular}
\end{table}

\begin{figure}[!t]
  \centering
  \includegraphics[width=\linewidth]{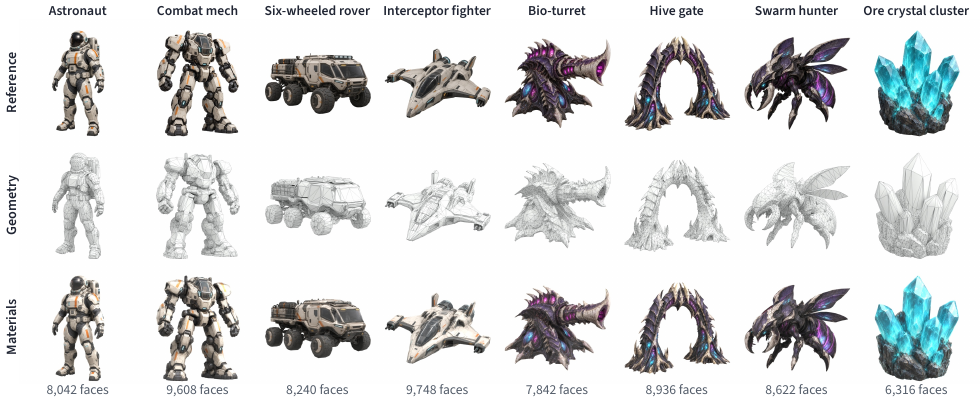}
  \caption{\textbf{Eight examples from the library of 123 assets.} The rows show reference images, generated geometry and the same meshes with externally refined materials. The mesh renders share the same camera, crop and scale. }
  \label{fig:assets}
\end{figure}

\subsection{Assets, assemblies and scenes}
\label{sec:assets}\label{sec:scenes}
\begin{figure}[!t]
  \centering
  \includegraphics[width=\linewidth]{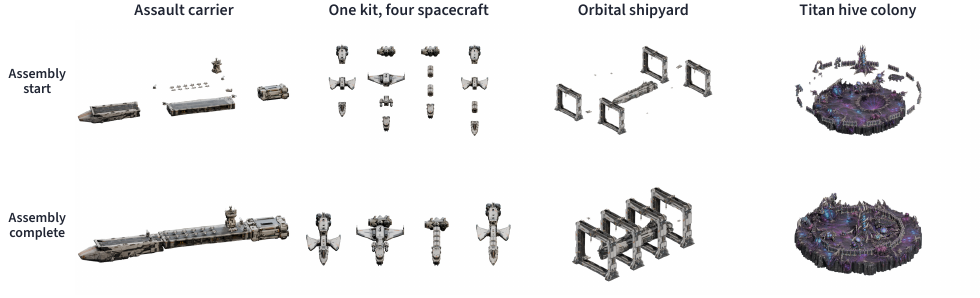}
  \caption{\textbf{Module reuse through assembly programs.} The columns show a carrier, four spacecraft built from one kit, an orbital shipyard and a Titan hive colony built from 39 instances of 17 asset types. The top and bottom rows show the initial and final configurations of each assembly. Only instance transforms change. Each pair shares the same view direction and lighting, with each foreground centered and scaled independently for display.}
  \label{fig:assemblies}

  \par\addvspace{\floatsep}
  \nextfloat
  \includegraphics[width=\linewidth]{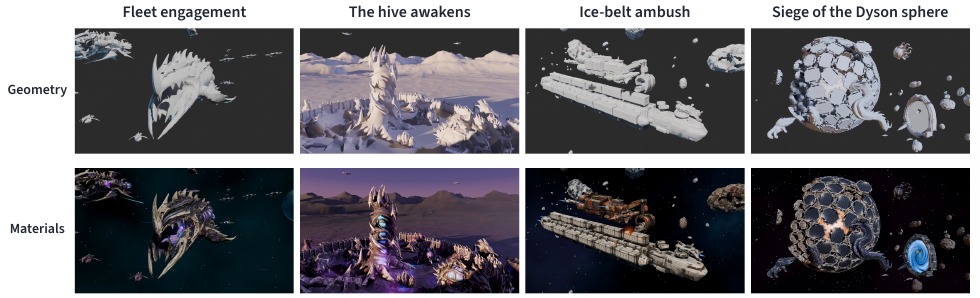}
  \caption{\textbf{Scenes composed by scene programs.} From left to right, the scenes are \emph{Fleet engagement}, \emph{The hive awakens}, \emph{Ice-belt ambush} and \emph{Siege of the Dyson sphere}. The top and bottom rows show engine renders of geometry and materials from the same scene state, frame and camera.}
  \label{fig:applications}
\end{figure}

Figure~\ref{fig:assets} shows the reference images, geometry and externally refined materials of eight of the 123 library assets.
The library spans mechanical, organic and environmental forms. We generated the eight library examples with MBR8. The interceptor fighter uses an earlier vertex checkpoint. The assets are reused in 19 assemblies and 12 animated scenes.
An assembly program places modules using instance transforms that preserve their geometry and materials.
In Figure~\ref{fig:assemblies}, one kit supplies modules for four spacecraft.
Scene programs add terrain, motion and cameras. Figure~\ref{fig:applications} pairs geometry and material renders of one frame from each of four scenes.

\subsection{Downstream agent effort}
\label{sec:agent}
We compare low-polygon and dense inputs for export under size and triangle limits, skeleton binding and flight, and rigid assembly.
In every task, the agent receives files and compact tool summaries.
The agent is \texttt{gpt-6.1-sol} at medium reasoning. The measurements exclude setup, orchestration and offline replay.
We compare a low-polygon mesh generated by MeshAxiom-Base with the native TRELLIS.2~\cite{xiang2025trellis2} output.
For budget export, the agent exports our 4,459-face mesh directly as a geometry-only GLB. The native TRELLIS.2 mesh requires simplification or reconstruction to meet the file size and triangle limits.

{\widowpenalty=10000
Table~\ref{tab:agent} reports the agent's token usage and elapsed time for the three tasks.
The low-polygon export runs use 65\% fewer output tokens, 9\% fewer uncached input and output tokens combined, and 76\% less time than the dense-input runs.
For rigging, the low-polygon input uses 11\% fewer output tokens and 2\% fewer uncached input and output tokens combined. For assembly, the corresponding reductions are 4\% and 32\%.\par}The same skinning, topology and motion checks take 0.138\,s on our mesh and 17.287\,s on native TRELLIS.2. Assembly replay takes 1.119\,s on our mesh and 1.570\,s on the high-poly one.

\begin{table}[!t]
  \centering
  \tablestyle{4pt}{1.1}
  \caption{\textbf{Agent effort with low-polygon and dense input geometry.} Both token columns give mean usage by the agent's language model. Each geometry was tested twice. Time covers the full agent run. The High poly rows use native TRELLIS.2~\cite{xiang2025trellis2}.}
  \label{tab:agent}
  \begin{adjustbox}{max width=\linewidth}
  \begin{tabular}{llrrr}
    \toprule
    Task & Geometry & Output tokens & \makecell{Uncached input\\+ output tokens} & Time (s) \\
    \midrule
    Budget export & Low poly (Ours) & 1,982 & 114,724 & 86.5 \\
     & High poly & 5,625 & 126,415 & 357.8 \\

    Rig and flight & Low poly (Ours) & 12,255 & 149,287 & 499.8 \\
     & High poly & 13,741 & 152,168 & 535.9 \\
    Rigid assembly & Low poly (Ours) & 2,177 & 116,476 & 91.9 \\
     & High poly & 2,262 & 172,417 & 106.8 \\
    \bottomrule
  \end{tabular}
  \end{adjustbox}
\end{table}

\FloatBarrier

\begin{table}[t]
  \centering
  \tablestyle{6pt}{1.1}
  \caption{\textbf{Execution cost per successful asset in the RSI pilot.} For each workflow, we divide total tokens and elapsed time across all attempts by the number of successful attempts. Tokens include uncached input tokens and all output tokens. Time combines workflow replay with the recorded duration of generation calls. These costs exclude learning the workflow.}
  \label{tab:rsi-success-cost}
  \begin{adjustbox}{max width=\linewidth}
  \begin{tabular}{lrr}
    \toprule
    Workflow & \makecell{Mean tokens\\per successful asset $\downarrow$} & \makecell{Mean time (s)\\per successful asset $\downarrow$} \\
    \midrule
    Original agent & 77,690 & 158.8 \\
    Re-run, no update & 69,006 & 151.6 \\
    Re-run, RSI (Ours) & 20,736 & 104.2 \\
    \bottomrule
  \end{tabular}
  \end{adjustbox}
\end{table}

\subsection{Pilot study: learning workflow skills}
\label{sec:rsi-pilot}
We selected 48 additional UE assets for this pilot, using 24 for skill learning and 24 for evaluation. Each workflow ran twice on each evaluation asset.
On each attempt, the original agent chose how to process, check, refine and submit the mesh. Both reruns in Table~\ref{tab:rsi-success-cost} started from the same workflow and traces, with model weights fixed. \emph{Re-run, no update} retained the original workflow, and \emph{Re-run, RSI (Ours)} used the learned workflow.
Under the same geometry, appearance and export requirements, the learned workflow used 73\% fewer tokens and took 34\% less time per successful asset than the original agent.

\section{Conclusion}
We presented MeshAxiom, a stack that generates explicit low-polygon meshes from individual images for use in programmable 3D worlds.
Its generator, MeshAxiom-Base, builds on the Nexus framework and takes the face budget as an input. At a budget of 4K faces, median generation time is 6.7\,s.
On 500 development objects, it obtains lower Chamfer, Hausdorff and edge distances than the 13 methods we compare with, including three dense generators after their official simplification.
Tools for checking, repair and materials, together with reusable asset records, carry the generated meshes into assemblies and animated scenes, which we demonstrated with 123 assets, 19 assemblies and 12 scenes.
In our export, rigging and assembly tests, the agent used fewer tokens and took less time with low-polygon inputs. In the RSI pilot, the workflow learned from episode traces also reduced both costs per successful asset.

\bibliographystyle{abbrv}
{\hbadness=2000 \bibliography{references}}
\end{document}